\documentclass[runningheads]{llncs}
\usepackage[T1]{fontenc}

\usepackage{graphicx}
\usepackage{amsmath}
\usepackage{amsfonts}
\usepackage{booktabs} 
\usepackage{multirow}
\usepackage{caption}
\usepackage{array}
\usepackage{makecell}
\begin{document}
\title{Curriculum-Based Multi-Tier Semantic Exploration via Deep Reinforcement Learning}
\titlerunning{Curriculum-Based Semantic Exploration via Deep Reinforcement Learning}
%
\author{
Abdel Hakim Drid\inst{1} \orcidID{0009-0004-8225-8968} \and
Vincenzo Suriani\inst{2} \orcidID{0000-0003-1199-8358} \and
Daniele Nardi\inst{3} \orcidID{0000-0001-6606-200X} \and
Abderrezzak Debilou\inst{1}  \orcidID{0009-0006-7523-8410}
}
\authorrunning{Drid et al.}
%
\institute{Department of Electrical Engineering - Mohamed Khider, University of Biskra, Biskra (Algeria) \\
\email{abdelhakim.drid@univ-biskra.dz \\ abderrazak.debilou@univ-biskra.dz}\\
\and
Department of Engineering - University of Basilicata, Potenza (Italy)
\email{vincenzo.suriani@unibas.it}\\
\and
Department of Computer, Control, and Management Engineering ``Antonio Ruberti'', Sapienza University of Rome, Rome (Italy)\\
\email{nardi@diag.uniroma1.it}}
\maketitle              
\begin{abstract}
Navigating and understanding complex and unknown environments autonomously demands more than just basic perception and movement from embodied agents. Truly effective exploration requires agents to possess higher-level cognitive abilities, the ability to reason about their surroundings, and make more informed decisions regarding exploration strategies. 
However, traditional RL approaches struggle to balance efficient exploration and semantic understanding due to limited cognitive capabilities embedded in the small policies for the agents, leading often to human drivers when dealing with semantic exploration.
In this paper, we address this challenge by presenting a novel Deep Reinforcement Learning (DRL) architecture that is specifically designed for resource efficient semantic exploration. A key methodological contribution is the integration of a Vision-Language Model (VLM) common-sense through a layered reward function. The VLM query is modeled as a dedicated action, allowing the agent to strategically query the VLM only when deemed necessary for gaining external guidance, thereby conserving resources. This mechanism is combined with a curriculum learning strategy designed to guide learning at different levels of complexity to ensure robust and stable learning. 
Our experimental evaluation results convincingly demonstrate that our agent achieves significantly enhanced object discovery rates and develops a learned capability to effectively navigate towards semantically rich regions. Furthermore, it also shows a strategic mastery of when to prompt for external environmental information. By demonstrating a practical and scalable method for embedding common-sense semantic reasoning with autonomous agents, this research provides a novel approach to pursuing a fully intelligent and self-guided exploration in robotics.

\end{abstract}
\section{Introduction} \label{sec:intro}
Exploration agents reconstructing maps of unknown environments face a unique challenge compared to goal-driven tasks like object navigation \cite{dorbala2023can}, \cite{ramakrishnan2022poni}, \cite{majumdar2022zson}. Goal-oriented agents benefit from explicit objectives that provide clear direction and feedback for learning \cite{chen2023not}. In contrast, exploration agents lack such external guidance and must intrinsically decide what constitutes valuable information and how to acquire it. This complexity is particularly relevant in map reconstruction, where the objective is to create a complete and accurate representation of an unknown environment.


While numerous techniques exist for map reconstruction, ranging from traditional graph-based methods to more recent learning-based approaches, many rely on human operators for guidance, intervention, or data annotation \cite{shaheer2023graph,sucar2021imap,singh2023efficient}. Traditional methods, particularly in SLAM, often require manual parameter tuning for optimal performance and may need human assistance for loop closure detection and map correction. Even some learning-based approaches may depend on labeled data for training or require human intervention to refine map representations or resolve ambiguities.
\begin{figure}[t]
    \centering
    \includegraphics[width=\textwidth, height = 5cm]{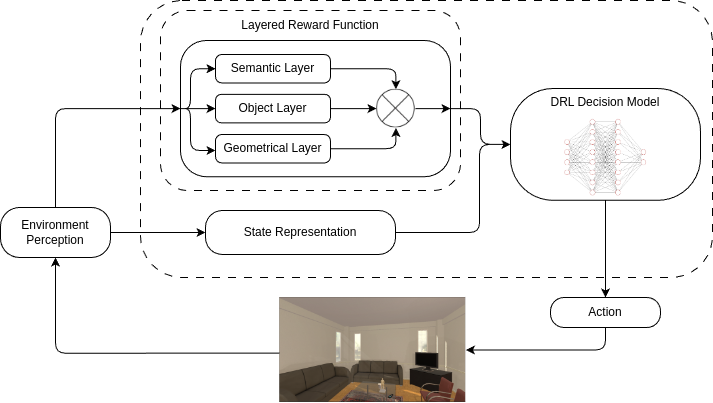}
    \caption{Architecture of our novel DRL agent, highlighting the key contribution: a "Layered Reward Function" consisting of "Geometrical Layer," "Objects Layer," and "Semantic Layer." This layered reward guides the agent's exploration policy, learned by the DRL Decision Model, to achieve semantically meaningful autonomous exploration.}
    \label{general_overview}
\end{figure}

Recently, vision-Language Models (VLMs) like CLIP \cite{radford2021learning}, GLIP \cite{li2022grounded}, LLaVA \cite{liu2023visual}, and BLIP \cite{li2022blip} have demonstrated significant potential for complex tasks by integrating visual and linguistic data. Effectively bridging the gap between perception, reasoning, and acting, these models allow agents to understand their surroundings with improved human-like comprehension \cite{liu2024vision}, \cite{argenziano2024empower}. This integration of common sense via VLMs facilitates better navigation, enhanced information inference, and more informed decision-making, making VLMs an ideal tool for exploration agents in unknown environments.

Fully realizing the potential of VLMs within DRL frameworks for autonomous exploration presents significant challenges. A key difficulty lies in designing reward functions that effectively translate rich VLM outputs into actionable guidance for the agent. The core challenge is not merely integrating VLMs and DRL, but creating an intricate reward mechanism that captures semantic information from the VLM and translates it into a coherent and efficient exploration strategy \cite{ma2024explorllm}, \cite{qu2024choices}. Moreover, exploration's inherent complexity, balancing spatial coverage, object interaction, and high-level semantic reasoning – makes it difficult to formulate a reward function that is simultaneously nuanced, informative, and stable.

To this end, we present a novel curriculum learning-based exploration agent with a layered reward function highlighted in (Fig. \ref{general_overview}) for autonomous navigation and semantically-informed mapping in unknown environments. The agent uses a three-layered reward system: geometrical (for exploration), object detection (guiding optimal viewpoints for object discovery), and semantic (via a VLM, for common-sense scene understanding). 
Key aspects of our method contributing to these results include:
\begin{itemize}
    \item Our curriculum learning strategy ensures stable and efficient integration of complex reward signals by progressively developing exploration skills.
    \item By integrating the VLM action-conditionally, our agent learns to strategically request external guidance, optimizing resource usage for maximal semantic information gain.
\end{itemize}

\section{Related Works} \label{sec:related}
Autonomous exploration and map reconstruction in an unknown environment without external human guidance is a core robotics challenge. Therefore, in this section, we examine the prior works that has been conducted to tackle this problem. In this section 
, we discuss the broader usage of Language Models (LMs) in Robotics navigation. Following that, 
 we narrow our focus to VLMs for Exploration.

The recent emergence and rapid advancements of LMs \cite{touvron2023llama}, \cite{liu2019roberta}, \cite{brown2020language}, \cite{team2024gemini}, \cite{grattafiori2024llama}, \cite{alayrac2022flamingo}, \cite{jia2021scaling} have sparked significant interest across various fields, including robotics. 
In particular, Visual Language Models can provide a better understanding of the agent' surrounding, bridging the perceptual input with language understanding, enabling the agent to leverage human-like semantic reasoning. Approaches like CLIP lead to application on a variety of robotic tasks, from exploration [CIT] to object grasping \cite{tziafas2023language}. 
In the exploration and navigation task, in \cite{wu2024voronav} the authors present VoroNav, a semantic exploration framework exploring a Reduced Voronoi Graph to strategically plan paths on unknown environments, enabling the agent to semantically reason through language models. 
Furthermore, VLMaps leverages VLM integrated with LLM to map environments spatially and interpret natural language commands into navigational instructions, enhancing spatial understanding without labeled data and enabling the agent to interpret complex spatial commands such as "in between the sofa and the TV" \cite{huang2023visual}. 
Additionally, several works have focused on enhancing the DRL model via pretrained semantic models. These approaches leverage embeddings from general-purpose VLMs to initialize policies, enabling agents to perform long-horizon tasks more efficiently by embedding semantic reasoning capabilities directly into their policy learning process \cite{chen2024vision}.
Addressing navigation safety, other studies specifically integrate collision avoidance mechanisms within semantic navigation frameworks, improving navigation safety and reducing collisions significantly in complex environments \cite{yue2024safe}.
Moreover, the field of object goal navigation has benefited from the introduction of sophisticated architectures like the Heterogeneous Zone Graph Visual Transformer, which effectively models spatial relationships and semantic interactions within environments. This approach enhances the agent's capability to efficiently navigate towards semantically relevant areas using graph-based reasoning \cite{he2024relation}. in comparison, our proposed approach introduces a distinctive DRL framework that instead of focusing on safety navigation or on semantic navigation, combines three levels of spatial understanding, balancing exploration capabilities and semantic depth. By combining a layered reward mechanism and a curriculum learning strategy, our methods optimize the use of Visual language models, empowering the agent with an action that asks for the VLM understanding, while embedding general common sense exploration capability in the learned policy.


\section{Methodology}
To address the challenges outlined in previous sections, we propose a novel VLM-DRL framework centered around a layered reward function and a curriculum learning strategy. 

\begin{figure}[t]
    \centering
    \includegraphics[width=\textwidth]{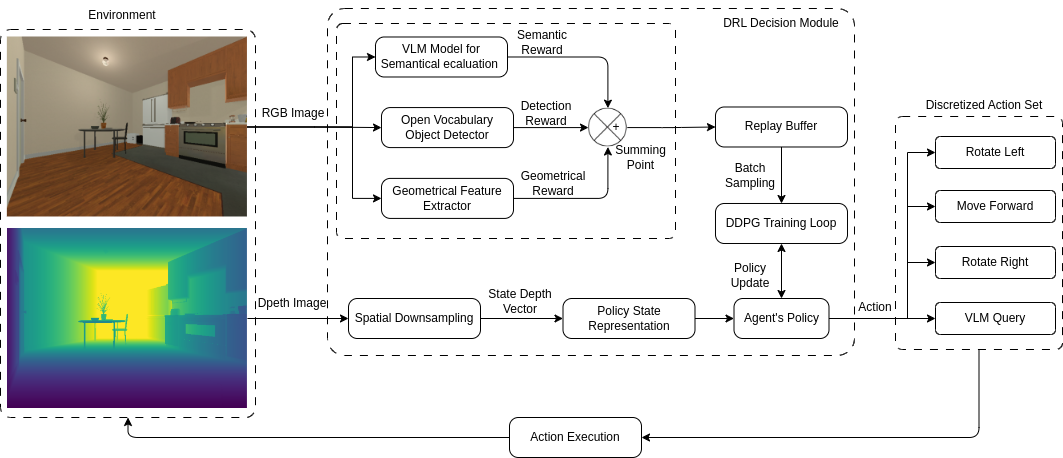}
    \caption{Architecture of our novel DRL agent, highlighting the key contribution: a "Layered Reward Function" consisting of "Geometrical Layer," "Objects Layer," and "Semantic Layer." This layered reward guides the agent's exploration policy, learned by the DRL Decision Model, to achieve semantically meaningful autonomous exploration.}
    \label{general_overview}
\end{figure}

\subsection{Reward Design}
In DRL-based exploration, where agents learn without predefined targets, a carefully designed reward function is crucial. This function must incentivize discovering new information while maintaining learning efficiency. To this end, we propose a hierarchical curiosity-driven reward mechanism to enhance exploration. This designed reward is explicitly designed to drive and encourage the agent's intrinsic curiosity, thereby enhancing exploration and guiding it towards meaningful states and a deeper understanding of the environment. 
%
%
\begin{equation} \label{Reward_EQ}
    \text{reward} = 
    \begin{cases} 
        R_C & \text{if $collision$} \\
        R_E & \text{otherwise}
    \end{cases}
\end{equation}
The proposed reward function is shown in (\ref{Reward_EQ}) where:
\begin{itemize}
    \item $R_C$ denotes the penalty when a collision occurs.
    \item $R_E$ denotes the exploration reward at each time step.
\end{itemize}
The $R_E$ reward is designed to guide the agent toward efficient exploration of the environment. When implementing $R_E$, it is crucial to include the most impactful information so that the agent not only explores effectively, but also progresses toward learning the optimal policy. 
To achieve this, $R_E$ is structured as a hierarchical mechanism, composed of several reward layers that contribute to different aspects of the agent’s exploration and learning process. These layers are outlined in Eq. (\ref{reward_eq})
\begin{equation} \label{reward_eq}
    R_E = \alpha * r_t^{(\mathrm{geom})} + \beta * r_t^{(\mathrm{obj})} + \delta * r_t^{(\mathrm{sematnical})}
\end{equation}
where each layer is defined as:
\begin{itemize} 
    \item $r_t^{(\mathrm{geom})}$ (Geometrical Features-Based Layer): Rewards exploration of new areas based on scene geometry and spatial layout.
    \item $r_t^{(\mathrm{obj})}$ (Object Detection Layer):  Encourages detecting more objects to encourage discovery of environmental elements.
    \item $r_t^{(\mathrm{semantnical})}$ (Semantic Layer): Guides exploration towards information-rich regions with high navigational potential based on semantic content.
    \item $\alpha$, $\beta$ and $\delta$: Reward component weights for balancing contributions.
\end{itemize}
%

These components are formulated to balance exploration and exploitation, guiding the agent toward discovering high-value states while maintaining efficiency in navigation.

\subsubsection{Geometric Features-Based Reward Layer:}
This component is based on geometrical features extracted from the scene frame to help the agent understand the geometry of the scene, encouraging exploration and movement through different areas.

We maintain a binary feature map matrix $feature\_map[x, y]$, where the $feature\_map[x, y] \in \{0, 1\}$ with dimensions equivalent to the captured agent frame. Initially, all entries are zero. Let $Geom\_features_t = \{ (x_i, y_i)\}_{\substack{i=1}}^{\mathrm{N_t}}$ denote the set of detected features key-points at time step $t$, with $N_t = |Geom\_features_t|$.

At each time step $t$ we process the scene frame to detect the features key-points. For each unique key-point $(x_i, y_i)$ in $Geom\_features_t$ the reward is calculated only if $feature\_map[x_i, y_i] = 0$:
\begin{itemize}
    \item We increment the count of newly discovered features, $C_t$ by one: $C_t \leftarrow C_t + 1$
    \item We update the matrix to mark the feature detected $feature\_map[x_i, y_i] \leftarrow 1$
\end{itemize}

Otherwise, the feature at this position has been detected in a previous step and $C_t$ is not incremented. 

The cumulative count of unique features up to the time step $t$ can be expressed as: $C = \sum_{t=1}^{T} C_t$ where $T$ is the current time step.

\subsubsection{Object Detection Reward Layer:}
We designed an object detection reward to encourage the discovery of novel environment aspects by specifically rewarding the agent for identifying newly detected object classes. Leveraging YOLO-World \cite{cheng2024yolo}, an open-vocabulary object detector, we maintain a memory of object classes detected within the current episode. This object detection reward, $r_t^{(\mathrm{obj})}$ is formulated as shown in (\ref{Eq:obj_detection}).
\begin{equation} \label{Eq:obj_detection}
    r_t^{(\mathrm{obj})} = min(N_{new\_objects}, N_{max\_objects})
\end{equation}
where:
\begin{itemize}
    \item $N_{new\_objects}$: Represents the novel detected objects at time-step $t$
    \item $N_{max\_objects}$: Hyperparameter capping the maximum number of new detected objects
\end{itemize}
By rewarding the discovery of new object classes, capped at $N_{max\_objects}$ to ensure balanced learning, this layer guides the agent towards exploring regions rich in diverse objects, contributing to a more detailed and informative environmental representation.
\subsubsection{Semantic Reward Layer (VLM-Driven):}
We integrate common-sense scene understanding into our exploration agent through a semantic reward layer driven by GPT-4o \cite{achiam2023gpt}. This advanced VLM, recognized for its multimodal reasoning and strong visual-language task performance, evaluates the desirability of different scenes for exploration based on a specific semantic prompt, making the reward action-dependent.
To enhance efficiency, instead of prompting the VLM at every time-step, our agent's action space includes a specific "VLM-Query" action. When the agent selects this action, the VLM is prompted with the current RGB image $I_t$ and a semantic evaluation prompt. The VLM returns a semantic score ($SC(I_t) \in [-1.0, +1.0]$), which is then discretized into three levels to generate a stable semantic reward $r_t^{(\mathrm{semantnical})}$.
\begin{equation} \label{Eq:semantic_reward}
    \text{$r_t^{(\mathrm{semantnical})}$} = 
    \begin{cases} 
        -1.0 & \text{if $-1.0 \leq SC < -0.3$} \\
        0.0 & \text{if $-0.3 \leq SC < +0.3$} \\
        +1.0 & \text{if $+0.3 \leq SC < +1.0$}
    \end{cases}
\end{equation}
The semantic reward $r_t^{(\mathrm{semantnical})}$ is formulated as in (\ref{Eq:semantic_reward}). To discourage frequent "VLM-Query" actions, we penalize consecutive uses. By making VLM evaluation an explicit action, we enable the agent to learn when it is most beneficial to incur the cost of querying the VLM, leading to a more strategic approach to semantically-informed exploration.

Collectively, these hierarchically structured reward components encourage an emergent exploratory curiosity behavior within the agent, compelling it to actively seek out and prioritize unexplored regions. By intrinsically rewarding the discovery of novel visual features, new objects, and ultimately, VLM-identified information-rich scenes. Such an integrated, curiosity-driven reward system, while distinct in its use of semantic guidance, shares a foundational goal of intrinsic motivation with methods like ICM \cite{pathak2017curiosity}, and is key to efficiently building a comprehensive understanding of the unknown environment.
%
\subsection{Decision Module}
With the layered reward function defined to guide semantic exploration, we now detail the Decision Module. This DRL-based module learns an optimal policy to maximize these rewards, ultimately enabling effective environmental exploration.

For efficient and robust exploration policy learning, we chose the Deep Deterministic Policy Gradient (DDPG) algorithm \cite{lillicrap2015continuous} for our Decision Model. As an off-policy method, DDPG provides sample efficiency, which is advantageous due to the computational demands of VLM queries and exploration episodes. The actor-critic structure of DDPG further promotes stable and effective policy learning.

Within the DDPG framework, the critic network learns to estimate the Q-value based on the summed reward signal (as shown in (\ref{Eq:critic_loss})) from our layered reward function. This comprehensive reward signal enables a full evaluation of state-action pairs. Then, the actor network develops a deterministic policy aimed at maximizing these Q-value estimates, effectively guiding the agent toward actions that fulfill the combined objectives of the layered reward.
\begin{equation} \label{Eq:critic_loss}
    L(\theta^Q) = \mathbb{E} [ (r_t + \gamma Q( s_{t+1}, \pi(s_{t+1} | \theta^\pi) | \theta^Q)  - Q(s_t,a_t | \theta^Q) )^2 ]
\end{equation}
In this context, $r_t$ is the reward received following the action $a_t$ taken in state $s_t$ under the network parameters $\theta$, and $\gamma$ is the discount factor.

\begin{figure}[t]
    \centering
    \includegraphics[width=\textwidth, height = 3cm]{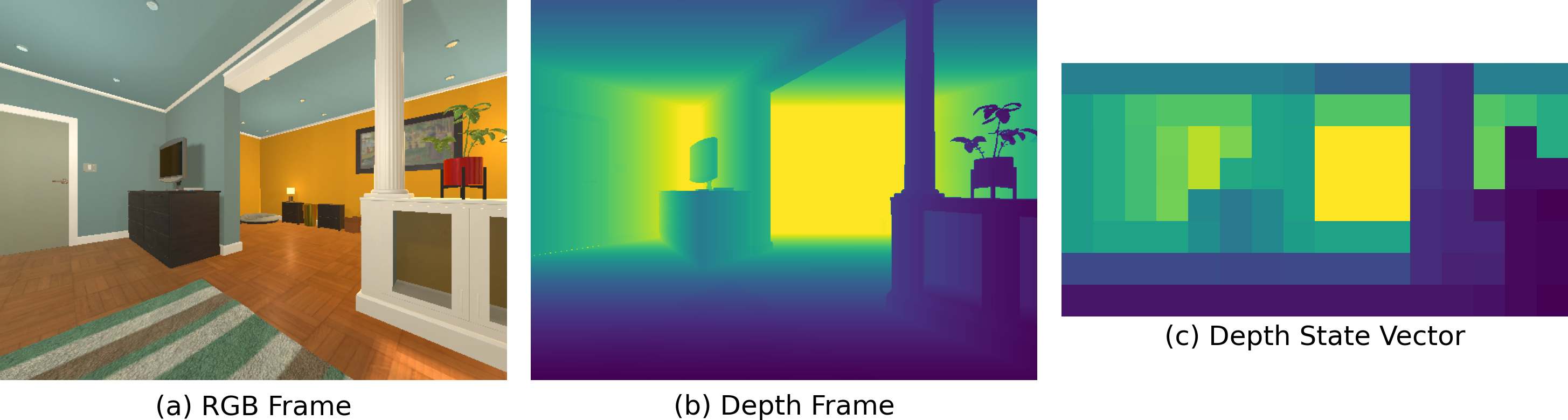}
    \caption{Concise state representation designed for efficient learning. (a) RGB Frame. (b) Corresponding Depth Frame. (c) Resulting 128-dimensional Depth State Vector after spatial downsampling.}
    \label{fig:state_representation}
\end{figure}

To equip the agent with local spatial awareness for exploration, we developed a geometrical state representation, as showed in Figure (\ref{fig:state_representation}). Utilizing RGB (Figure (\ref{fig:state_representation})(a)) and Depth (Figure (\ref{fig:state_representation})(b)) frames, we spatially downsample the 480x640 Depth Frame to generate a 128-dimensional Depth State Vector (Figure (\ref{fig:state_representation})(c)). This vector constitutes the agent's state representation.


\paragraph{Action Space:}
For simplified and robust control, we discretize the exploration agent's action space. While the DDPG actor provides a continuous action output, we convert this into a set of four discrete actions for execution. This approach simplifies implementation and improves robustness, while still leveraging DDPG's continuous action learning. The policy's action $a_{\pi}$ is drawn from the discrete action set:

$\mathcal{A}_{discrete} = \{ \text{''RotateLeft''}, \text{''MoveForward''}, \text{''RotateRight''}, \text{''VLM-Query''} \}$
\subsection{Curriculum Training Methodology}
Having detailed the design of our layered reward function and the architecture of our decision model, we now turn to the Training Methodology that enables effective learning of our semantic exploration agent. 
To facilitate robust and efficient learning, we employed a curriculum learning strategy, dividing the training process into three distinct phases:
\paragraph{{Phase 1 (Geometrical Exploration)}}: Phase one utilizes solely the geometric reward layer to establish basic navigation skills and environmental awareness. This focuses the agent on broad exploration, rewarding forward movement, and obstacle avoidance. 
\paragraph{Phase 2 (Object-Aware Exploration):} Phase 2 builds on Phase 1 navigation skills by adding the Object Detection Reward Layer alongside the Geometric Reward Layer. The object detection reward guides the agent to object-rich areas, encouraging "object-awareness" and enriching environmental understanding.
%
%
\paragraph{Phase 3 (Semantic Exploration):} Phase 3, the final stage, enables information-seeking exploration using common-sense semantics. The agent learns strategic VLM queries for deeper scene understanding, leading to a policy that actively seeks semantic input and prioritizes exploration of informative areas.
%
%
\section{Experimental Setup}
After detailing the fundamentals of our approach in the previous sections, in this we will detail the conducted simulation experiments and the obtained results.  
\paragraph{Simulation Environment}: The training experiments of our semantic exploration agent were performed within the AI2-THOR interactive environment \cite{kolve2017ai2}. AI2-THOR is a widely used simulation platform for embodied AI agent research, offering a rich set of photorealistic 3D indoor scenes, detailed object models, and a physics engine capable of simulating realistic agent-environment interactions. AI2-THOR environments offer realistic indoor layouts (e.g., kitchens, bedrooms), dense interactive objects (furniture, appliances), and varied scene configurations, creating challenging exploration scenarios. 
\paragraph{Scenes Details and Layout}: AI2-THOR offers 120 scenes spread over 4 scenarios (Kitchens, Bedrooms, Living rooms, and Bathrooms). From this wide range of scenarios, we select a subset that captures a range of difficulties, scene layouts, and spatial object arrangements. This selection allows for a more comprehensive evaluation of the agent's exploration capabilities across different environment layouts. Furthermore, to encourage generalization and exploration strategies, at the start of each training episode, the agent's initial state is randomized (position and orientation) within the selected scene.
%
%
%
\section{Experimental Results}
We now present the results obtained from training and evaluating our proposed semantic exploration agent. This section systematically presents our quantitative findings, ablation studies, and qualitative observations to assess the effectiveness of our approach. 
To accelerate DRL training and VLM inference, experiments were performed using an NVIDIA GeForce RTX 3060 GPU and a 
Intel Core i5-10400F CPU leveraging on the CUDA library.
\subsection{Quantitative Findings}
For each phase, we have tested the agent's performance on a variety of AI2-THOR scenes. Table \ref{tab:overall_info} focuses on three key quantitative metrics: 
\begin{itemize}
    \item Maximum Path Length(Max PL): Indicating the extent of agent movement
    \item Total Number of Detected Objects(TDO): Reflecting object discovery
    \item Total Confidence Scores (TCS): Representing the cumulative sum of object detection confidence scores
\end{itemize}
While Max PL shows a slight decrease in later phases, both TDO and TCS exhibit substantial increases as the curriculum progresses, demonstrating the impact of later reward layers. Metrics are marked as not applicable ('/') for Phase 1 as those rewards are not active in this initial phase.
\begin{table}[t]
    \centering
    \small
    \caption{Overall Quantitative Performance on 30 Test Scenes achieved by our agent, averaged across the test scenes}
    \begin{tabular}{  >{\centering\arraybackslash} m{2cm} | >{\centering\arraybackslash} m{2cm} | >{\centering\arraybackslash} m{2cm} | >{\centering\arraybackslash} m{2cm} }
    \hline
     \textbf{Phase}  & \textbf{Max PL} & \textbf{TDO} & \textbf{TCS} \\ \hline
     Phase 1 & 8.75 &   /  &    /   \\ \hline
     Phase 2 & 5.75 & 1254 & 485.09 \\ \hline
     Phase 3 &   5  & 1274 & 500.09 \\ \hline
    \end{tabular}
    \label{tab:overall_info}
\end{table}
In Table \ref{tab:choosen_paths} we present the same metrics but specifically for four chosen scenes that are visually analyzed in detail later in this section. This table provides the quantitative data for visually presented scenarios.
\begin{table}[t]
    \centering
    \caption{Scene-Specific Quantitative Results for Four Illustrative Scenes, 
             presented individually for four chosen AI2-THOR test scenes}
    \small 
    \begin{tabular}{c|c|c|c|c|c|c|c|c|c|c|c|c}
        \toprule
        \multirow{3}{*}{Scene} 
        & \multicolumn{4}{c|}{Geometrical Layer} 
        & \multicolumn{4}{c|}{Detection Layer} 
        & \multicolumn{4}{c}{Semantic Layer} \\
        \cmidrule(lr){2-5}\cmidrule(lr){6-9}\cmidrule(lr){10-13}
        & \multicolumn{4}{c|}{$\alpha = 1.0, \;\beta = 0.0, \;\delta = 0.0$}
        & \multicolumn{4}{c|}{$\alpha = 0.25, \;\beta = 0.75, \;\delta = 0.0$}
        & \multicolumn{4}{c}{$\alpha = 0.25, \;\beta = 0.75, \;\delta = 2.0$}\\
        \cmidrule(lr){2-5}\cmidrule(lr){6-9}\cmidrule(lr){10-13}
         & PL & TDO & TCS & MC 
         & PL & TDO & TCS & MC 
         & PL & TDO & TCS & MC \\
        \midrule
        Scene 1 & 7.00 & / & / & / & 1.25 & 45 & 16.48 & 0.37 & 5.00 & 48 & 17.72 & 0.39 \\
        Scene 2 & 8.25 & / & / & / & 4.75 & 45 & 17.07 & 0.38 & 3.25 & 63 & 24.9  & 0.40 \\
        Scene 3 & 4.25 & / & / & / & 1.25 & 29 & 11.69 & 0.40 & 2.00 & 46 & 18.87 & 0.41 \\
        Scene 4 & 6.75 & / & / & / & 0.25 & 31 & 13.64 & 0.38 & 1.75 & 45 & 17.00 & 0.44 \\
        \bottomrule
    \end{tabular}
    \label{tab:choosen_paths}
\end{table}
Notably, within each scene, observe the consistent increase in TDO and TCS as we progress from the Geometrical Layer to the Semantic Layer configuration, suggesting a progressive benefit of adding object detection and semantic guidance.
\subsection{Qualitative Findings:}
\begin{figure}[!b]
    \centering
    \includegraphics[width = \textwidth, height = 6cm]{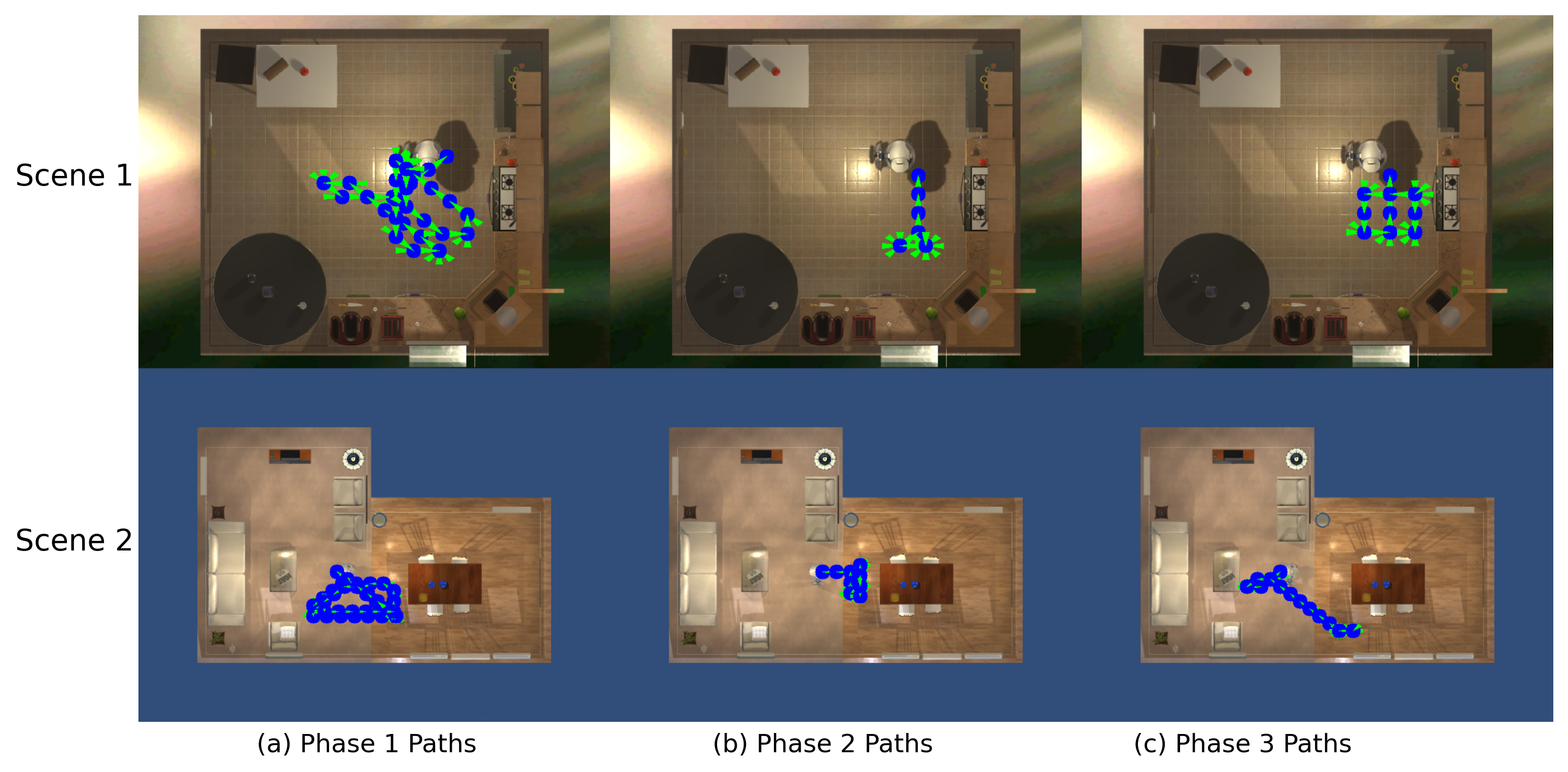}
    \caption{Agent Trajectories Across Four Illustrative AI2-THOR Scenes. Scene 1 - 4 visualizes trajectories in different environments. Each column represents a different curriculum learning phase of our semantic exploration agent.}
    \label{fig:stacked_paths}
\end{figure}

To visually showcase our agent's exploration, Fig. (\ref{fig:stacked_paths}) presents example trajectories in AI2-THOR scenes. Each scene displays trajectories learned after successive reward layers from our curriculum, illustrating evolving exploratory patterns. These paths demonstrate how the agent increasingly improves at navigating complex layouts, prioritizing informative areas, and avoiding redundant revisits as it incorporates more sophisticated reward signals. Such qualitative visualizations offer direct insight into the agent's maturing, intelligent environmental interaction.
%
\subsection{Ablation Study}
To assess individual component contributions, we performed ablation studies, detailed in Table \ref{tab:ablation}. The table presents performance metrics (Max PL, TDO, and Total Detector Calls (TDC)) for various input shape and reward Type combinations, including configurations with and without an additional "Accuracy Reward" component (acc) that considers the detection accuracy of objects. 

This format allows direct analysis of these parameters' impact on the exploration performance.
\begin{table}[t]
    \centering
    \small
    \caption{Ablation Study Results: Impact of Reward Type.}
    \begin{tabular}{| c | c | c | c | c | c |}
    \hline
     \textbf{Input Shape} & \textbf{Reward Type} & \textbf{Max PL} & \textbf{TDO} & \textbf{TDC} & \textbf{Info}\\ \hline
     37  & geom + obj        & 3.5    & 180 & 84  & Few AI2-THOR scenes \\ \hline
     73  & geom + obj        & 2.978  & 201 & 88  & Increased state \\ \hline
     37  & geom + obj + acc  & 3.10   & 212 & 108 & Added accuracy component \\ \hline
     73  & geom + obj + acc  & 5.991  & 535 & 228 & \makecell{Increased state/scenes with \\  domain randomization} \\ \hline
     128 & geom + obj        & 12.748 & 621 & 250 & Increased state vector \\ \hline
    \end{tabular}
    \label{tab:ablation}
\end{table}
In Figure (\ref{fig:ablation}), we present bar plots visualizing the performance metrics for each ablation experiment. The figure shows separate subplots for Max PL, TDO, and TDC, allowing for a direct visual comparison of how different experimental configurations impact each metric.
\section{Discussion}
This paper introduces a novel architecture for autonomous exploration, specifically designed to leverage the power of VLMs and DRL for semantically informed navigation and environmental understanding. The experimental results presented demonstrate that our proposed approach effectively addresses the challenges of exploration in unknown environments without pre-defined objectives. Notably, our agent exhibits enhanced object discovery and semantically guided exploration behaviors. 
The quantitative results in Table \ref{tab:overall_info} show that both Total TDO and TCS exhibit a marked and progressive increase from Phase 1 to Phase 3. This suggests that the agent becomes significantly more effective at discovering objects and understanding the semantic content of the environment as it progresses through the curriculum.

Analyzing quantitative results in Table \ref{tab:choosen_paths} highlights a clear shift in exploration strategy across phases. Phase 1's higher Max PL indicates broader, geometry-driven spatial exploration. However, later phases show slightly reduced Max PL but a high number of both TDO and TCS, in Phases 2 and 3. This performance is further demonstrated in Fig. (\ref{fig:stacked_paths}), where the object detection reward guides the agent path towards more object-aware behavior, while adding the semantic layer further pushes the agent towards a more efficient, information-prioritizing exploration approach, rather than simply maximizing distance traveled in case of phase 1.

\begin{figure}[t]
    \centering
    \includegraphics[width=\textwidth, height = 3cm]{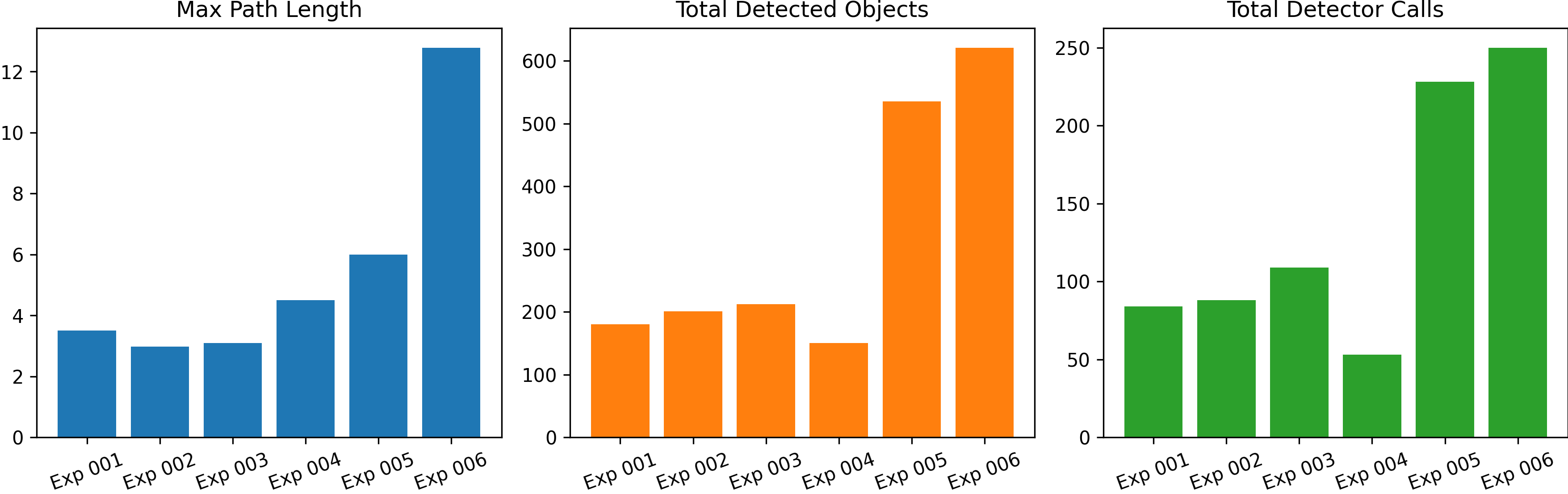}
    \caption{This figure presents bar plots for three key performance metrics across different ablation experiments. Each subplot shows the metric values for Experiment IDs.}
    \label{fig:ablation}
\end{figure}

Ablation studies from Table \ref{tab:ablation} revealed that reward component additions, such as accuracy, did not substantially enhance exploration performance. Interestingly, even with a simplified expert assistance mechanism (detector calls replacing "VLM-Query" for computational efficiency), agents still increased 'detector calls' in complex environments, as also shown in Fig. (\ref{fig:ablation}). This indicates a generalizable tendency to seek external expert guidance when facing more challenging exploration tasks.

\section{Conclusion}
In this paper, we have tackled the challenge of autonomous exploration in unknown environments by implementing a novel DRL architecture. Our core lies in integrating a VLM within a layered reward function and curriculum learning paradigm. Experimental results demonstrate that our approach effectively leverages VLM for common-sense scene understanding, resulting in agents capable of enhanced object discovery and semantically guided navigation. This work represents a significant step towards creating fully autonomous exploration agents that can learn and operate effectively in complex, unstructured environments without human intervention.
Future work will focus on extending this architecture to real-world robotic platforms and further investigating the strategic utilization of external knowledge sources for even more sophisticated exploration.

\bibliographystyle{splncs04}
\bibliography{references}
\end{document}